\def\year{2020}\relax
\documentclass[letterpaper]{article} 
\usepackage{aaai20}  
\usepackage{times}  
\usepackage{helvet} 
\usepackage{courier}  
\usepackage[hyphens]{url}  
\usepackage{graphicx} 
\usepackage{graphicx}  
\usepackage[ruled,vlined]{algorithm2e}
\usepackage{amsfonts}
\usepackage{amsmath}
\title{Transfer learning based few-shot classification using optimal transport mapping from preprocessed latent space of backbone neural network}
\author{Tom\'{a}\v{s} Chobola, Daniel Va\v{s}ata and Pavel Kord\'{i}k\\ 
Faculty of Information Technology, Czech Technical University in Prague\\ 
Thakurova 9\\
Prague, Czech Republic\\
choboto1@fit.cvut.cz 
}
\begin{document}
\maketitle
\begin{abstract}
MetaDL Challenge 2020 focused on image classification tasks in few-shot settings. This paper describes second best submission in the competition. Our meta learning approach modifies the distribution of classes in a latent space produced by a backbone network for each class in order to better follow the Gaussian distribution. After this operation which we call Latent Space Transform algorithm, centers of classes are further aligned in an iterative fashion of the Expectation Maximisation algorithm to utilize information in unlabeled data that are often provided on top of few labelled instances. For this task, we utilize optimal transport mapping using the Sinkhorn algorithm. Our experiments show that this approach outperforms previous works as well as other variants of the algorithm, using K-Nearest Neighbour algorithm, Gaussian Mixture Models, etc.
\end{abstract}

\section{Introduction}
Few-shot learning is increasingly popular because it can handle machine learning tasks with just a few learning examples. It is also more biologically plausible and closer to what we observe in nature. While learning a new task, one normally does not start from a randomly initialised neural network presenting hundreds of thousands of examples in several thousands epochs. 

When you are told to remember a person from a picture, you are able to distinguish this person from others even when you see her in different positions or environments. In machine learning, this is called one shot learning. The task of one shot learning is to learn new classes given only one instance available for each class. Three-way five-shot learning means learning three classes given five training instances each. You do not learn classifiers from scratch, but you typically use neural networks trained on similar tasks using much more data. This also reflects the natural situation when the visual perception is already well trained on similar tasks when trying to remember a new person from the picture. 
This process can be also called meta learning or transfer learning as one uses a pretrained neural network called a backbone network. 
Also, in a few-shot learning scenario, you can often utilise unlabelled instances apart of those few labelled samples that are available for the task.

MetaDl challenge 2020\footnote{\url{https://competitions.codalab.org/competitions/26638}} focused on few shot learning of image classification tasks. Participants trained a meta-learner on a meta-train set and produced a learner which was subsequently used to train on classification tasks generated from the meta-test set and evaluated. The goal was to discover learners with ability to quickly adapt to new unseen image classification tasks. 

Our submissions scored second in the final leaderboard. This paper describes methods we have experimented with and the architecture of the meta-learning pipeline responsible for second best result in the competition.
The architecture of our solution mainly follows \cite{hu2020leveraging} with important improvements in the preprocessing of latent space output of the backbone model $B$. The main improvement is in the different normalization of the transformed feature vectors which resembles the Gaussian distribution assumption better. 
Since this is the key assumption for the proper functionality of the Sinkhorn mapping algorithm, it leads to more accurate results.

\section{Related Work}
There are several different approaches to few shot learning. The survey \cite{wang2020generalizing} is a good resource to learn about general overview and taxonomy of few shot learning methods. 
Prototypical networks \cite{snell2017prototypical} and the Siamese networks \cite{koch2015siamese} focus on learning embeddings transforming the data in a way that it can be recognised with a simple classifier. This approach is further enhanced by relation networks \cite{sung2018learning} which is able to classify images of new classes by predicting distances between query images and the few examples of each new class.

Another interesting direction aims at the learning process itself. In \cite{DBLP:conf/iclr/RaviL17} a recurrent network based meta-learner model learns the exact optimization algorithm used to train
another learner neural network classifier in the few-shot setup. Meta-transfer learning \cite{sun2019meta} adapts a deep neural network for few shot learning tasks. Transfer is achieved by learning scaling and shifting functions of DNN weights for each task.

\begin{figure*}[h]
  \includegraphics[width=\linewidth]{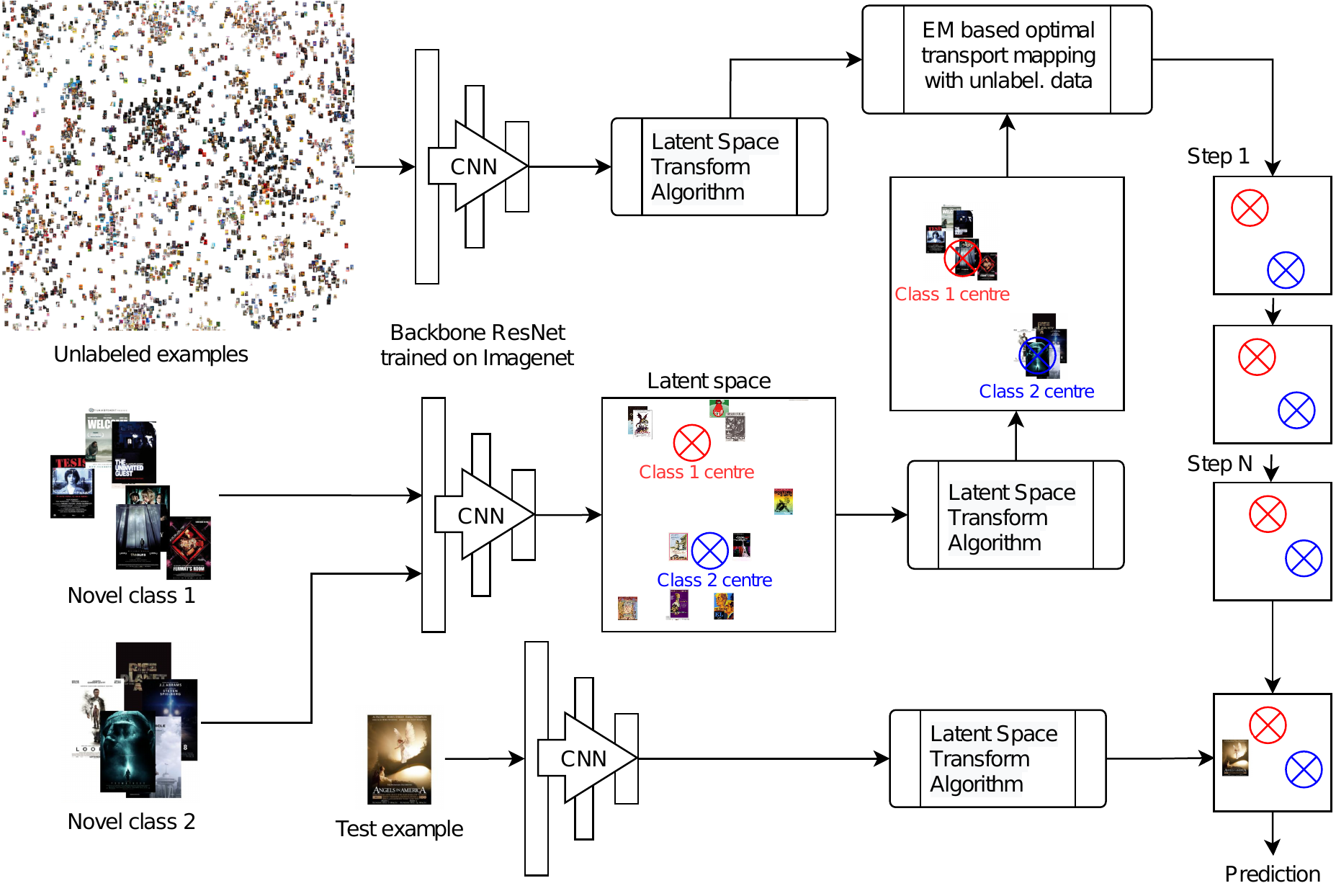}
  \caption{In order to predict the class label of a test example, we transform the image using a backbone CNN to the latent space and preprocess vectors by the Latent Space Transform algorithm that helps to transform distribution of individual classes to Gaussian like. Then a test example is processed and compared to the class centres that have been iteratively adjusted using a Sinkhorn mapping with unlabeled data projected to the latent space in the same way. The closest class is assigned to the test example as prediction.}
  \label{fig:diagram}
\end{figure*}

We further extend the direction of few-shot learning research that is leveraging classification capabilities in robust backbone models (neural networks) pretrained on similar tasks. These {\bf transfer learning based methods} need to find mapping of few-shot classes to similar classes used to train the backbone model. 

In \cite{rohrbach2013transfer} the Propagated Semantic Transfer has been applied to employ semantic knowledge transfer to original classes, combine the transferred predictions with labels for the novel classes, exploit the manifold structure of novel classes by graph based learning and improve the local neighborhood in such graph structures by replacing the raw feature-based representation with an attribute-based representation. 

When transferring the knowledge, deep embeddings are far superior, compared to weight transfer, as a starting point for novel tasks as investigated in \cite{scott2018adapted}. Another similar approach is TransMatch \cite{yu2020transmatch}, where a feature extractor is pre-trained on original classes and subsequently used to initialize few-shot classifier weights for the novel classes, the classifier is also updated with a semisupervised learning method. 

Our research proceeds from \cite{hu2020leveraging}, where the latent space produced by a backbone deep network is preprocessed by a power transform and optimal-transport algorithm maps original classes to novel classes while centres on new classes are iteratively adjusted. This approach has shown significant improvement in accuracy in our experiments. The importance of feature transformation for few-shot learning is confirmed by \cite{wang2019simpleshot}. 

\section{Model description}
Formally, in a few-shot learning task one has a dataset $D$ containing a part $D_S$ with a few labelled samples from $w$ classes and 
a part $D_Q$ with some unlabelled samples.
The goal is to predict the classes for samples in $D_Q$. We will assume that $D_S$ contains exactly $s$ labelled samples for each class and $D_Q$ contains 
exactly $q$ unlabelled samples for each class. Hence, there are $w s$ samples in $D_S$ and $w q$ samples in $D_Q$. The $i$-th sample from $D$ will be denoted by 
$x_i$ and if it is from $D_S$ we will denote its label by $y_i$.

Moreover, let us assume that there is another dataset $D_B$ corresponding to some related task, such as image classification to some novel classes.
This dataset can be used to train the backbone model $b$ which maps the initial space into some latent feature space $\mathcal L = \mathbb R^d$. 
In order to train such a model one might train the neural network for classification and then remove the last classification layers as we did in the experiments. 
Or an encoder part of an autoencoder might be used. 

The next step is to preprocess the points in the latent space to be prepared for the final prediction algorithm that estimates the labels.
As was recently researched this step is crucial and may lead to significant improvements of the result, see \cite{wang2019simpleshot}.
To proceed we will further assume that the features obtained from the backbone model $B$ are non-negative, i.e. $\mathcal L = \mathbb R_+^d$. 
This is often the case when one extracts $b$ as a part of some neural network with the ReLU activation function on inner layers. 
Let us denote by $B$ the dataset $D$ transformed by $b$ and by $B_S$ and $B_Q$ its parts corresponding to $D_S$ and $D_Q$, respectively.

In the preprocessing we transform the dataset $D$ of points in the latent space $\mathcal L$ to a final dataset $F$ of points in the final 
feature space $\mathcal F = \mathbb R^r$, where the dimension 
$r = \min\{d, w (s + q)\}$ is the minimum of the dimension $d$ of $\mathcal L$ and the number of points in the dataset $D$.
The preprocessing is a composition of three steps and we will call it the Latent Space Transform algorithm (LST). The first is the power transform combined with the semi-normalization of each point given by
\[
    f_1(u) = \frac{(u + \varepsilon)^{\beta}}{\|(u + \varepsilon)^\beta\|_{2}^{\delta}}\quad \mathrm{for\ all}\ u \in \mathcal L,
\]
where the power is taken component-wise, $\varepsilon = 10^{-6}$ is the normalization parameter, and $\| \cdot \|$ is the Euclidean norm.
The hyperparameter $\beta$ controls the strength of the power transform and the hyperparameter $\delta$ controls the strength of 
the normalization, where $\delta = 1$ means the full normalization and $\delta = 0$ yields no normalization at all.
The power transform is known to help stabilising the variance and making the data more Gaussian distribution-like by reducing its skewness, see \cite{boxcox1964}.
The normalization on the other hand leads to the projection on the unit sphere which is not compatible with the assumption used later in the optimal-transport that the components of points in the same class are independent with Gaussian distribution of the same variance.
Hence, the semi-normalization controlled by the hyperparameter $\delta$ enables for having some variance in the perpendicular 
direction to the unit sphere surface and thus does not a priori break the compatibility of the resulting distribution with the Gaussian assumption.
Let us denote the dataset with all points in $B$ transformed using $f_1$ by $F_1$ and $F_{1,S}, F_{1,Q}$ analogously.

The second step is the removal of unnecessary dimensions using the QR decomposition of the transposition of the already preprocessed data matrix 
$\mathbf{F}_{1} \in \mathbb R^{w(s+q), d}$ corresponding to dataset $F_1$,
\[
    \mathbf F_1^T = \mathbf Q \mathbf R
\]
and thus we define
\[
    \mathbf F_2 = \mathbf F_1 \mathbf Q
\]
so that $\mathbf F_2 \in \mathbb R^{w(s+q), r}$, where $r = \min\{d, w (s + q)\}$, and the corresponding dataset is denoted by $F_2$.
We again denote by $F_{2,S}$ and $F_{2,Q}$ the parts of $F_2$ that corresponds to samples originally in $D_S$ and $D_Q$, respectively.
It corresponds to the change of the orthonormal basis in the $\mathbb R^d$ and throwing away the dimensions that are zero for the data points.

\begin{figure}[h]
  \includegraphics[width=\linewidth]{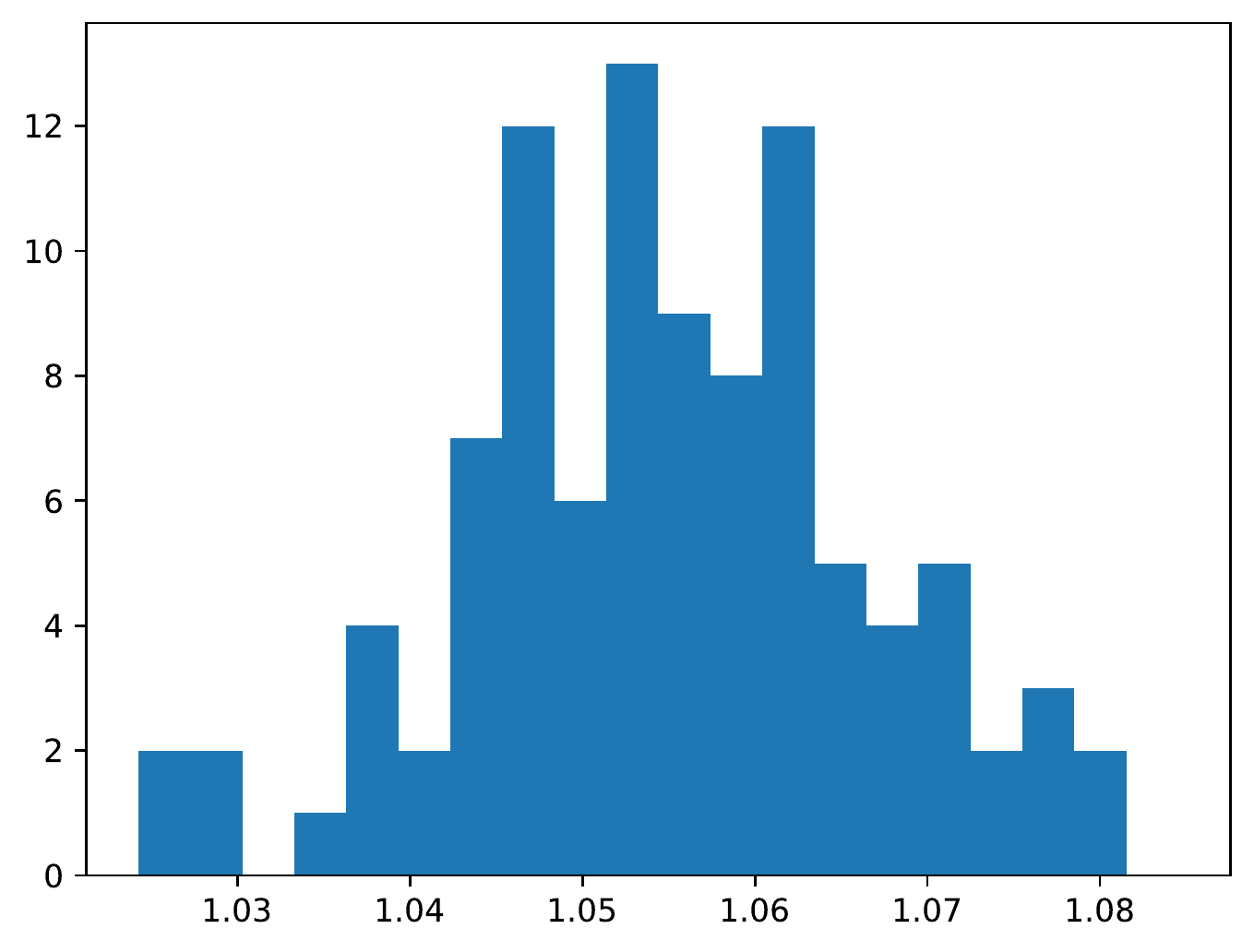}
  \caption{Latent Space Transform algorithm produces Gaussian like distribution also for the norms of the transformed samples. 
  The figure was produced for one batch from the CUB dataset with $s = 5, q = 15, \beta = 0.5, \delta = 0.3$, and $\gamma = 0.9$.}
  \label{fig:normsLST}
\end{figure}

The last preprocessing step is the centering and further semi-normalization given by
\[
    f_3(u) = \frac{u - \bar u}{\|u\|_{2}^{\gamma}},
\]
where 
\[
    \bar u = \frac{1}{w(s+q)}\sum_{i = 1}^{w(s+q)} u_i
\]
is the centroid (component-wise average) of the dataset $F_2$.
Again, the hyperparameter $\gamma$ allows to control the strength of the normalization. For $\gamma < 1$ the resulting points are only partially
normalized and one may expect to better resemble the Gaussian distribution assumed in the next step. The typical result for the final Euclidean norms of transformed points is shown in Figure \ref{fig:normsLST}.

Let us denote the final preprocessed dataset by $F$ and its respective parts corresponding to original parts $D_S$ and $D_Q$ by $F_S$ and $F_Q$, respectively.

Once the preprocessing of the dataset is finished, the actual optimal-transport can begin. In this part we directly follow \cite{hu2020leveraging}.
The preliminary assumption of the method is the independent Gaussian distributions of all components of points in individual classes with class centres $c_1, \dots, c_w$ as parameters. Moreover, it is assumed that all the Gaussian distributions have the same variance $\lambda/2$, where $\lambda$ is the hyperparameter.
Under this assumption the maximum a posteriori estimate (MAP) $\hat y_1, \dots, \hat y_{wq}$ of 
the labels of unlabelled samples $f_1, \dots, f_{wq}$ from $F_Q$ corresponds to
\begin{multline*}
    \{\hat y_j\}_{j=1}^{wq}, \{\hat c_k\}_{k=1}^w =  \arg\max_{\{y_j\}, \{c_k\}} \prod_i P(y_i | f_i) \\
    = \arg\max_{\{y_j\}, \{c_k\}} \prod_i P(f_i | y_i) P(y_i) \\
    = \arg\max_{\{y_j\}, \{c_k\}} \prod_i e^{-\lambda^{-1} \|f_i - c_{y_i}\|_2^2} P(y_i).
\end{multline*}
This is directly related to the Optimal Transport theory, see \cite{hu2020leveraging,marco2013Sinkhorn,berman2020,villani2003},
and one may use the iterative expectation-maximization like approach incorporating the Sinkhorn algorithm to get the MAP estimate.
It consists of repeating of two steps, where the first is the construction of the mapping matrix $\mathbf M^*$ with elements $\mathbf M^*_{ij} = P(y_i = j)$ which is maximizing the previous term for a given centres $c_1, \dots, c_w$ and the second step is the estimation of class centres 
that is for the fixed 
mapping matrix again optimizing the previous term.
For the Sinkhorn algorithm, see  \cite{marco2013Sinkhorn} the mapping matrix is defined as
\begin{multline*}
    \mathbf M^* = \mathrm{Sinkhorn}(\mathbf L,a,b,\lambda) \\
    = \arg\min_{\mathbf M\in U(a,b)}\sum_{i, j}\mathbf M_{ij} \mathbf L_{ij} + \lambda H(\mathbf M),
\end{multline*}
where $U(a,b)$ is a set of positive matrices in $\mathbb R^{wq\times w}$ for which the rows sums to a vector $a$ and columns sums to a vector $b$, 
$\mathbf L\in R^{wq\times w}$ is the cost function consisting of Euclidean distances between unlabelled instances and class centres, that is 
$\mathbf L_{ij} = \|f_i - c_{j}\|_2^2$, the hyperparameter $\lambda$ is a regularisation coefficient forcing the entropy 
$H(\mathbf M)=-\sum_{ij} \mathbf M_{ij}\log \mathbf M_{ij}$ to become smaller, 
$a$ denotes the distribution of the amount that each unlabelled example uses for class allocation, i.e. $a$ is the vector of ones with $wq$ elements, 
and $b$ denotes the distribution of the amount of unlabelled examples allocated to each class, i.e. $b$ is the vector with $w$ elements that equals to $q$.

The iterative approach starts with initialising the class centres from the labelled samples in $F_S$. 
Then the mapping matrix $\mathbf M^*$ is calculated using the Sinkhorn algorithm. It is then used to re-estimate the class centres via the update using
\[
    \mu_j=\frac{\sum_{f_i \in F_Q}\mathbf M^*_{ij}f_i+\sum_{f_k\in F_S, y_k = j}f_k}{s + \sum_{i=1}^{wq}\mathbf M^*_{ij}}.
\]
To avoid unnecessarily big steps in centre estimations, the new centre is set to be $c_j = c_j + \alpha(\mu_j-c_j)$, where the $\alpha$ is the learning rate. The number of iterations is fixed to $n_{\mathrm{steps}}$. Once the iteration process finishes, the labels of the samples from $F_Q$ might be estimated
from the last mapping matrix as 
\[
    \hat y_i = \arg\max_j \mathbf M^*_{ij}.
\]
The overview of the algorithm is given in Algorithm \ref{alg:sink}. The overall process of our approach is depicted in Figure \ref{fig:diagram}. The code is available at https://github.com/ctom2/latent-space-transform.

\SetKwFor{RepTimes}{repeat}{times:}{end}
\begin{algorithm}
\SetAlgoLined
\textbf{Parameters:} $w,s,q,\lambda,\alpha,n_{\mathrm{steps}}$\\
\textbf{Initialisation:} $c_j=\frac{1}{s}\sum_{f_k\in F_S, y_k = j}f_k$\\
 \RepTimes{$n_{\mathrm{steps}}$}{
  $\mathbf L_{ij}=||f_i-c_j||^2, \forall i,j$\;
  $\mathbf M^*=\mathrm{Sinkhorn}(\mathbf L,p=1_{wq},q=q1_w,\lambda)$\;
  Calculate $\mu_j$\;
  $c_j=c_j+\alpha(\mu_j-c_j)$\;
 }
 \Return{$\hat y_i = \arg\max_j \mathbf M^*_{ij}$}
 \caption{Optimal map algorithm}
 \label{alg:sink}
\end{algorithm}

\section{Experiments}
The performance of the stated methods was measured based on standardised few-shot classification datasets CIFAR-FS \cite{bertinetto2019metalearning} and CUB \cite{WahCUB_200_2011}. CIFAR-FS dataset consists of images with size of $32\times 32$ distributed into 100 classes, each containing 600 images. The dataset is split into 64 base classes, 16 validation classes and 20 novel classes. CUB dataset contains 11,788 images of birds, each with size $84\times 84$, distributed over 200 classes. The dataset is split into 100 base classes, 50 validation classes and 50 novel classes.

In each testing run, $w$ classes are randomly and uniformly drawn from novel classes, where each class consists of $s$ instances with label and $q$ instances without label.

\begin{table}[]
\caption{Hyperparameters used in the final evaluation of the LST+MAP model.}
\vspace{5mm}
\label{tab:hyperparams}
\centering
\begin{tabular}{c|cc|cc}
\hline
         & \multicolumn{2}{c|}{\textbf{1-shot}} & \multicolumn{2}{c}{\textbf{5-shot}} \\ \cline{2-5} 
Parameter & CIFAR-FS            & CUB            & CIFAR-FS            & CUB            \\ \hline
$\beta$ & 0.5                 & 0.5            & 0.5                 & 0.5            \\ \hline
$\lambda$ & 10                 & 10            & 10                 & 10            \\ \hline
$\alpha$ & 0.3                 & 0.4            & 0.2                 & 0.2            \\ \hline
$n_{\textit{steps}}$ & 20                 & 30            & 20                 & 20            \\ \hline
$\delta$ & 0.3                 & 0.7            & 0.4                 & 0.3            \\ \hline
$\gamma$ & 0.98                & 0.95           & 0.95                & 0.9  \\ \hline         
\end{tabular}
\end{table}

Because of the high performance of WideResNet \cite{zagoruyko2017wide} augmented with the S2M2 method \cite{mangla2020charting} in the few-shot setting, we chose it as the backbone architecture for our model. The latent representation of images produced by the backbone is a vector with dimension of 640. The QR decomposition reduces the said dimension to 80 in 1-shot setting, and to 100 in 5-shot setting.

All experiments are based on $w=5,q=15$ and $s=1$ or $5$. To evaluate the performance of the models we run 10,000 random draws to obtain mean accuracy with $95\%$ confidence scores. 

\begin{table}[h!]
\caption{1-shot accuracy of models based on Power Transform (PT), our proposed Latent Space Transform (LST) and WideResNet backbone.}
\vspace{5mm}
\label{tab:accuracies_1shot}
\centering
\resizebox{.99\columnwidth}{!}{%
\begin{tabular}{l|c|cc}
\hline
               &          & \multicolumn{2}{c}{\textbf{1-shot}}                                                             \\
Method         & Backbone & \multicolumn{1}{c}{CIFAR}                              & CUB                                   \\ \hline
PT+MAP         & WRN      & \multicolumn{1}{c}{$87.69 \pm 0.23\%$}                  & $91.55\pm 0.19\%$                     \\ \hline
PT+GMM          & WRN      & \multicolumn{1}{l}{$86.96\pm 0.22\%$} & \multicolumn{1}{l}{$90.06\pm 0.18\%$} \\ \hline
PT+KNN         & WRN      & \multicolumn{1}{c}{$86.17 \pm 0.19\%$}                  & $89.07\pm 0.17\%$ \\ \hline\hline    
LST+MAP         & WRN      & \multicolumn{1}{c}{$\mathbf{87.79 \pm 0.23\%}$}                  & $\mathbf{91.68\pm 0.19\%}$                     \\ \hline
LST+GMM         & WRN      & \multicolumn{1}{c}{$87.01\pm 0.21\%$}                  & $89.9 \pm 0.18\%$                     \\ \hline
LST+KNN         & WRN      & \multicolumn{1}{c}{$85.76\pm 0.19\%$}                  & $89.26 \pm 0.17\%$                     \\ \hline

\end{tabular}
}
\end{table}

By tuning the hyperparameters of the model we observed evolution in accuracy in both 1-shot and 5-shot setting with dependency on tested dataset. The overview with the hyperparameters can be found in Table \ref{tab:hyperparams}. The final accuracy can be seen in Table \ref{tab:accuracies_1shot} and Table \ref{tab:accuracies_5shot} for 1-shot and 5-shot setting, respectively. Moreover, the tables include results obtained by substituting MAP with different clustering algorithms, Gaussian Mixture model and $k$-means model, that take the transformed features as their input. The $k$-means model is initiated with centres corresponding to the labeled instances in a testing run. The centres are then iteratively refined to produce better representations of the class centres. Similarly, Gaussian Mixture model is provided with initial means corresponding to the labeled examples at the beginning of each run. To compare our proposed transform method with the Power Transform (PT) \cite{hu2020leveraging}, we performed the same substitutions for the PT+MAP model. 

The scores show that even by omitting the MAP part from the architecture and replacing it with simpler classification approaches while keeping the transformation intact produces competitive results. Moreover, to compare the statistical significance of the superiority of the LST+MAP model against the PT+MAP model we performed the paired t-test with $p$-values presented in Table \ref{tab:ttest}. We can see that except for the CUB dataset in 5-shot scenario the LST+MAP model is significantly better than the PT+MAP model.

In terms of execution time, we measured an average of $0.0026s$ per run in 1-shot setting and $0.003s$ per run in 5-shot setting with the GPU backend.

\begin{table}[h!]
\caption{5-shot accuracy of models based on Power Transform (PT), our proposed Latent Space Transform (LST) and WideResNet backbone. The authors of the PT+MAP model presented accuracy $93.99\pm 0.10\%$ in 5-shot setting for CUB dataset, however we were able to obtain higher accuracy with their described model configuration.}
\vspace{5mm}
\label{tab:accuracies_5shot}
\centering
\resizebox{.99\columnwidth}{!}{%
\begin{tabular}{l|c|cc}
\hline
               &          & \multicolumn{2}{c}{\textbf{5-shot}}                                                             \\
Method         & Backbone & \multicolumn{1}{c}{CIFAR}                              & CUB                                   \\ \hline
PT+MAP         & WRN      & \multicolumn{1}{c}{$90.68\pm 0.15\%$}                  & $94.09\pm 0.09\%$                     \\ \hline
PT+GMM          & WRN      & \multicolumn{1}{l}{$87.16\pm 0.21\%$} & \multicolumn{1}{l}{$90.04\pm 0.20\%$} \\ \hline
PT+KNN         & WRN      & \multicolumn{1}{c}{$86.70\pm 0.19\%$}                  & $89.72\pm 0.18\%$ \\ \hline\hline    
LST+MAP         & WRN      & \multicolumn{1}{c}{$\mathbf{90.73\pm 0.15\%}$}                  & $\mathbf{94.09\pm 0.09\%}$                  \\ \hline
LST+GMM         & WRN      & \multicolumn{1}{c}{$87.33\pm 0.20\%$}                  & $90.06\pm 0.18\%$                     \\ \hline
LST+KNN         & WRN      & \multicolumn{1}{c}{$86.56\pm 0.18\%$}                  & $89.64\pm 0.18\%$                     \\ \hline

\end{tabular}
}
\end{table}

\begin{table}[]
\caption{$p$-values of the paired t-test with the null hypothesis that the accuracy of the PT+MAP model is greater or equal than the accuracy of the LST+MAP model against the alternative that the accuracy of the PT+MAP model is smaller than the accuracy of the LST+MAP model.}
\vspace{5mm}
\label{tab:ttest}
\centering
\resizebox{.99\columnwidth}{!}{%
\begin{tabular}{c|cc|cc}
\hline
            & \multicolumn{2}{c|}{\textbf{1-shot}} & \multicolumn{2}{c}{\textbf{5-shot}} \\ \cline{2-5} 
            & CIFAR-FS          & CUB              & CIFAR-FS          & CUB              \\ \hline
$p$-value   & $9.09\mathrm{e}{-5}$           & $1.99\mathrm{e}{-9}$          & $1.68\mathrm{e}{-7}$           & $0.78$    \\ \hline        
\end{tabular}
}
\end{table}

\section{Challenge submission}
In this section, we describe modification to our method we have elaborated for the MetaDl challenge 2020. The main limitation of the challenge was the submission runtime which had to include backbone training time and was limited to two hours. Therefore we were not able to utilise the WRN backbone as we suggest above.

Our best performing solution was relying on a lighter backbone network based on the ResNet architecture. During the backbone training, the fed images could either be left as they were, or their saturation or brightness could be changed with the probability set to $1/3$ for each alteration. Moreover, the training batches also included the same images rotated by 90, 180 and 270 degrees to further improve the backbone capabilities and augment the training overall. 

\section{Conclusion}
Extracted features from backbones often do not resemble Gaussian-like distributions, even though multiple algorithms are built on that assumption. In this paper we show how to transform feature vectors into better Gaussian-like distributions. By applying an iterative optimal-transport algorithm to estimate class centres empirically, the subsequent clustering method gains significant improvement over other few-shot classification methods. 

Our experiments confirmed that the Latent Space Transform algorithm introduced above outperforms other forms of feature preprocessing including the Power Transform. We have also compared our approach based on optimal transport mapping to other classification methods based on Gaussian mixtures and nearest neighbours. For both CIFAR and CUB datasets, our approach proved to be superior in both 1-shot and 5-shot learning scenarios. 

We have adjusted our method for the MetaDl challenge 2020 competition and scored second in the final leaderboard.

\section{Acknowledgment}
This work was supported by the Student Summer Research Program 2020 of FIT CTU in Prague. Moreover, the research was supported by the Grant Agency of the Czech Technical University in Prague (SGS20/213/OHK3/3T/18) and the Czech Science Foundation (GA\v{C}R 18-18080S).

\bibliographystyle{aaai}
\bibliography{references}

\end{document}